\documentclass[journal]{IEEEtran}
\usepackage{cite}
\usepackage{amssymb,amsfonts}
\usepackage{algorithmic}
\usepackage{graphicx}
\usepackage{textcomp}
\newcommand{\add}[1] {\textcolor{blue}{#1}} 

\usepackage[utf8]{inputenc} 
\usepackage[T1]{fontenc}    
\usepackage{hyperref}       
\usepackage{url}            
\usepackage{booktabs}       
\usepackage{amsfonts}       
\usepackage{nicefrac}       
\usepackage{microtype}      
\usepackage{xcolor}         
\usepackage[linesnumbered,ruled,vlined]{algorithm2e}

\usepackage{pifont}
\newcommand{\cmark}{\ding{51}}%
\newcommand{\xmark}{\ding{55}}%

\usepackage{amsmath}
\usepackage{amssymb}
\usepackage{latexsym}
\usepackage{graphicx}
\usepackage{multirow}
\usepackage{adjustbox}
\usepackage{makecell}

\usepackage{mathrsfs}

\def\BibTeX{{\rm B\kern-.05em{\sc i\kern-.025em b}\kern-.08em
    T\kern-.1667em\lower.7ex\hbox{E}\kern-.125emX}}

\begin{document}

\title{Single-round Self-supervised Distributed Learning using Vision Transformer}

\author{Sangjoon Park, Ik-jae Lee, Jun Won Kim, and Jong Chul Ye \IEEEmembership{Fellow, IEEE}
\thanks{Submitted: April 12, 2023}
\thanks{Sangjoon Park is with the Department of Bio and Brain Engineering, Korea Advanced Institute of Science and Technology (KAIST), Daejeon Republic of Korea and the Department of Radiation Oncology, Yonsei University College of Medicine, Seoul, Republic of Korea (E-mail: depecher@kaist.ac.kr).}
\thanks{Ik-Jae Lee and Jun Won Kim are with the Department of Radiation Oncology, Gangnam Severance Hospital, Seoul, Republic of Korea (E-mail: ikjae412@yuhs.ac, junwon@yuhs.ac).}
\thanks{Jong Chul Ye is with the Kim Jaechul Graduate School of AI, KAIST, Daejeon, Republic of Korea (E-mail: jong.ye@kaist.ac.kr).}
\thanks{Jong Chul Ye and Jun Won Kim are co-corresponding authors.}
}

\maketitle

\begin{abstract}

Despite the recent success of deep learning in the field of medicine, the issue of data scarcity is exacerbated by concerns about privacy and data ownership. Distributed learning approaches, including federated learning, have been investigated to address these issues. However, they are hindered by the need for cumbersome communication overheads and weaknesses in privacy protection. To tackle these challenges, we propose a self-supervised masked sampling distillation method for the vision transformer. This method can be implemented without continuous communication and can enhance privacy by utilizing a vision transformer-specific encryption technique. We conducted extensive experiments on two different tasks, which demonstrated the effectiveness of our method. We achieved superior performance compared to the existing distributed learning strategy as well as the fine-tuning only baseline. Furthermore, since the self-supervised model created using our proposed method can achieve a general semantic understanding of the image, we demonstrate its potential as a task-agnostic self-supervised foundation model for various downstream tasks, thereby expanding its applicability in the medical domain.

\end{abstract}


\section{Introduction}
\label{sec:introduction}

\IEEEPARstart{D}{eep} learning has established itself as the standard for developing artificial intelligence (AI)-powered medical tools. However, its reliance on data and labels necessitates collaboration among multiple institutions. Unfortunately, strict regulations often impede the free sharing of patient-derived data due to privacy concerns \cite{hoofnagle2019european, edemekong2018health}. To address this challenge, distributed learning methods such as federated learning (FL) \cite{konevcny2016federated} have been introduced to alleviate issues related to data governance and ownership. These methods enable the sharing of de-identified data between collaborators under formal consent, overcoming one of the most significant obstacles in AI research.

In FL, the objective is to train a model on the server-side while preserving the training data on multiple client-side edge devices. The global model is distributed by the central server to each client, which performs training iterations with their data in parallel. The local updates obtained are then sent back to the server, which conducts parallel training iterations using their local data. The resulting local updates are then transmitted back to the server, which aggregates and averages them before disseminating the updated global model. This iterative process continues until the model converges. Despite resolving data sharing issues, FL does not guarantee complete privacy, as private data can be compromised through inversion attacks that reconstruct private data by utilizing stolen gradients from insecure aggregation \cite{geiping2020inverting}. Furthermore, FL imposes significant computational loads on client-side edge devices, as most computations and model updates are performed locally \cite{li2020federated}. Additionally, the practical implementation of FL introduces considerable communication overheads, as the entire typically large-sized model needs to be aggregated and distributed between the server and clients.

Recently, the vision community has introduced Vision Transformer (ViT), a deep learning model that exclusively relies on attention mechanisms \cite{dosovitskiy2020image}. ViT's powerful yet simple attention architecture has made it an indispensable tool in vision research. Recent studies have revealed that ViT exhibits a shape-biased nature akin to that of humans and is less susceptible to perturbations such as occlusion or random patch permutation \cite{naseer2021intriguing}. Furthermore, various self-supervised learning (SSL) methods have been proposed for ViT, including knowledge distillation-based semantic meaning learning \cite{caron2021emerging} and masked image modeling \cite{bao2021beit,he2022masked}.


Building upon the valuable characteristics of ViT, we propose a novel approach for distributed SSL that facilitates the creation of a self-supervised model with only a single round of communication between the server and clients. Our approach utilizes the permutation invariant properties of self-attention to provide encryption via \textit{feature-space random permutation}, as proposed in a previous work \cite{9934926}. Furthermore, we leverage another important property of ViT, which stems from its patch-based image processing, to enable random masked sampling-based SSL to train the task-agnostic self-supervised model solely on the server-side. This enables single-round FL without requiring continuous interaction between the server and clients.

    
    

\section{Related Works}

\subsection{Self-supervised learning of ViT}

Caron et al. \cite{caron2021emerging} presented a groundbreaking contrastive learning method, called Distillation without a Label (DINO), for ViT that enables the model to learn the task-agnostic semantic meaning of images without the need for cumbersome negative samples as in traditional contrastive learning methods. This is accomplished through the use of teacher-student knowledge distillation and a random multi-crop strategy, which facilitates local-to-global correspondence learning. Attention visualization revealed that this method is especially effective for ViT.


The patch-based attention architecture of ViT has also been applied for SSL using the masked image modeling approach. This approach is akin to the masked language modeling approach utilized in pre-training Bidirectional Encoder Representations from Transformers  (BERT) for natural language processing. In their work, Bao et al. \cite{bao2021beit} presented the BERT, which leverages discrete visual tokens obtained through a discrete tokenizer to predict the masked patch. He et al. \cite{he2022masked}, on the other hand, proposed a simpler masked autoencoder (MAE) strategy that employs an efficient encoder-decoder design to directly predict pixels within the masked patches. It is important to note that these strategies are tailored for the patch-wise image processing of ViT and may not be suitable for CNN-based models that employ shared convolution kernels across the image.

In the medical imaging, some pioneering works have attempted to combine these SSL approaches of ViT with FL to address the challenges of label insufficiency. Wu et al. \cite{wu2022federated} performed self-supervised FL via contrastive learning or MAE using the unlabeled data on the client-side, and then fine-tuned with client-side label, demonstrating superior performance compared to random initialization or local SSL. Similarly, Yan et al. \cite{yan2023label} proposed a label-efficient self-supervised FL approach, where they divided the training process into two stages, utilizing the MAE approach for self-supervised FL on the local clients in the first stage, and performed supervised FL with labeled data in the second stage, thereby mitigating the issues of data heterogeneity and label insufficiency in FL.


\subsection{Federated Learning with ViT}

Several pioneering studies have investigated the effective use of ViT in distributed learning environments \cite{kim2022task, qu2022rethinking, park2021federated, 9934926}. Qu et al. \cite{qu2022rethinking} recently demonstrated that self-attention-based ViT architecture is more robust in FL among clients with heterogeneous data compared to CNN-based models.
The \textit{Federated Split Task-agnostic} (\textsc{FeSTA}) learning method was developed based on the modular structure of the ViT model, which consists of an embedder head, transformer body, and task-specific tail, combining the strengths of two different distributed learning methods, FL and split learning, to enhance the performance of individual tasks by facilitating collaboration between clients with different tasks \cite{park2021federated}. Recently, a new method called \textit{Federated Split Task-Agnostic Learning with Permutating Pure ViT (p-\textsc{FeSTA})} has been introduced to address the limitations of \textsc{FeSTA} \cite{9934926}. This method utilizes the permutation-invariant property of ViT by employing random patch permutation to enhance privacy and reduce communication overhead.
The primary goal of the \textit{p-\textsc{FeSTA}} method is to reduce communication between the server and clients and enhance privacy using a \textit{feature-space permutation module}. This is achieved by leveraging the permutation invariant property of the self-attention, which are the core components of ViT. The model is trained with permutated patch features in the feature space using the innovative \textit{feature-space permutation module}, thereby providing privacy protection and preventing malicious attackers from reconstructing private data from the intermediate features. The permutated features can be securely stored in the server-side memory and used throughout the entire learning process, reducing the burden by approximately 50\% compared to the original \textsc{FeSTA} method.
However, this method has certain limitations that restrict its general applicability. Firstly, continuous communication between the server and clients is required for model training, as labels are necessary to update the shared transformer body. Secondly, multi-task learning is only feasible among clients concurrently participating in distributed learning with relevant tasks. Lastly, as these methods are supervised learning approaches, manually labeled data is required, which can be hard to obtain in the medical domain.

\begin{figure}[!t]
\centering
\includegraphics[width=0.5\textwidth]{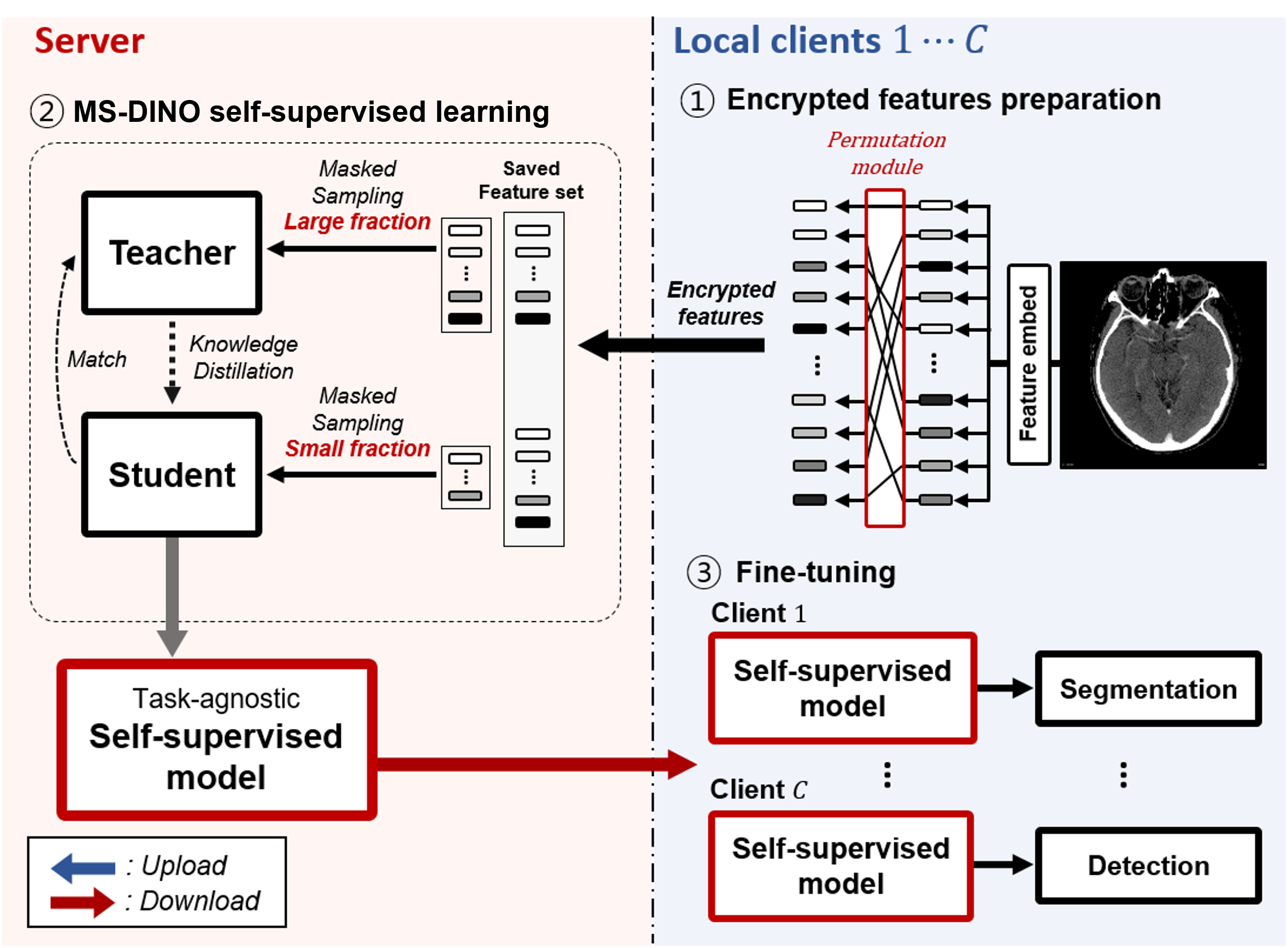}
\caption{The overall framework can be described as follows: (1) The participating clients utilize an arbitrary patch embedder and a \textit{feature-space permutation module} to encrypt the patch features prior to transmitting them to the server. (2) The server-side performs MS-DINO training to achieve an overall semantic understanding of the modality, ultimately producing a pre-trained model. (3) This pre-trained model is then accessible to authorized clients for various downstream tasks.
} 
\label{fig_framework}
\end{figure}

\begin{figure*}[!t]
\centering
\includegraphics[width=1.0\textwidth]{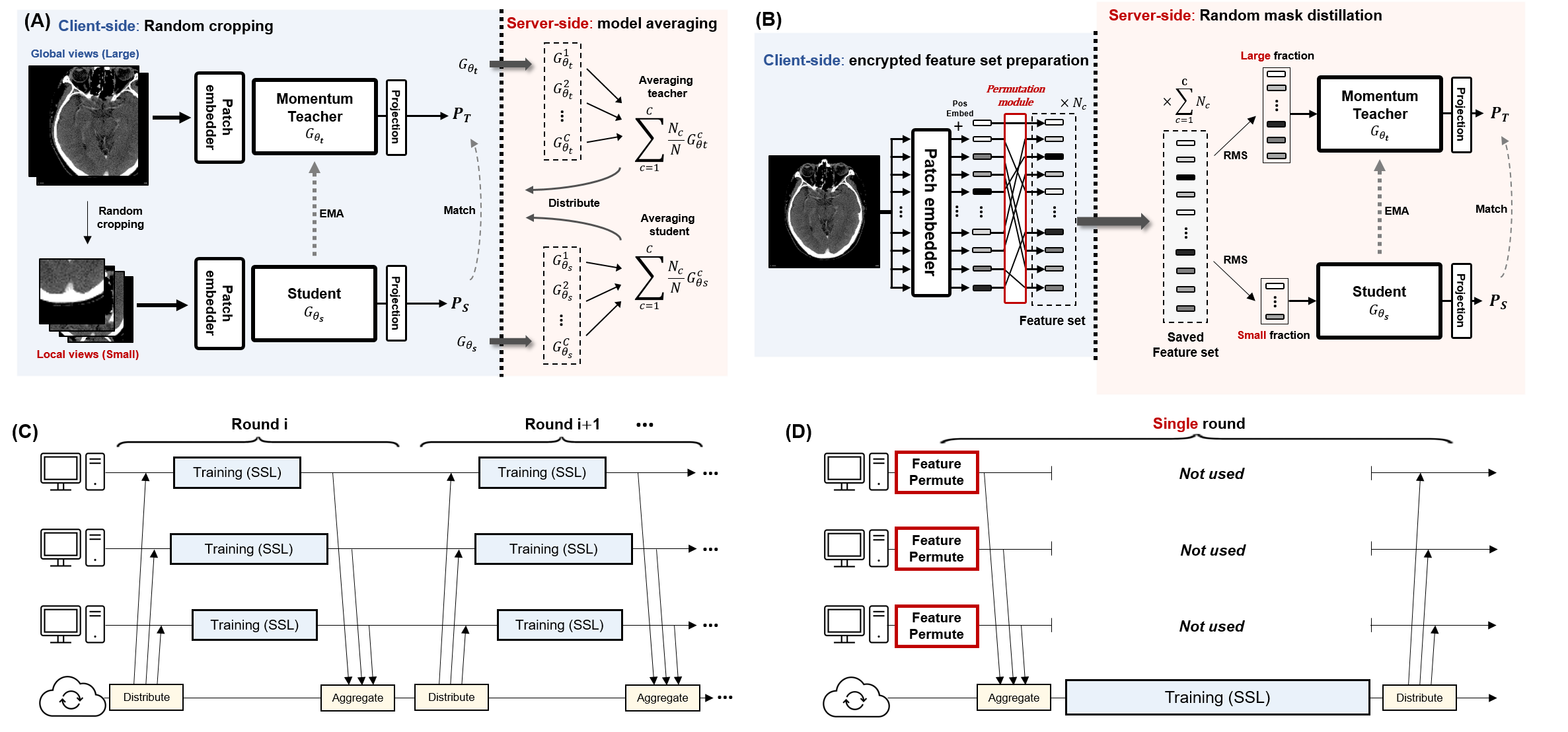}
\caption{The figure presented illustrates a comparison between the Distillation with No Labels (DINO) trained with the federated learning (FL) method and the Masked Sampling Distillation with No Labels (MS-DINO) method. (A) In the original DINO method, the model acquires a general semantic understanding of the image by way of local-to-global correspondence between the student and momentum teacher. (B) In contrast, the MS-DINO method teaches the model correspondence between a smaller and larger number by training with randomly permuted patch features. (C) During self-supervised learning (SSL) via FL, the server aggregates and distributes weight updates from each client every round, necessitating continuous communication. (D) In contrast, the proposed method utilizes a \textit{feature-space permutation module} to receive patch-permuted encrypted features on the server-side, subsequently employing these features throughout all SSL processes. As a result, training can be accomplished only with the single-round communication between the server and the clients. EMA, exponential moving average; RMS, random masked sampling.
} 
\label{fig_MS_DINO}
\end{figure*}

\section{Method}

We introduces the \textit{Masked Sampling Distillation with No Labels (MS-DINO)} method, which is designed for single-round distributed SSL. The method involves three main steps, as depicted in Figure~\ref{fig_framework}(A). Initially, an arbitrary patch embedder is employed to extract patch features from all data, along with an arbitrary position embedding. The features are randomly permuted using the \textit{\textit{feature-space permutation module}}. Subsequently, the resulting encrypted patch features are stored on the server-side, where random masked feature sampling-based SSL is performed independently, resulting in a task-agnostic self-supervised ViT model equipped with semantic feature extraction capabilities. Finally, authorized clients can access the pre-trained model to apply it to downstream tasks, leading to the improved performances compared with the baselines without SSL.


\subsection{Self-Supervised Learning}

The MS-DINO method's SSL process is illustrated in Figure~\ref{fig_MS_DINO}. The method involves extracting patch features $f$ with an arbitrary patch feature extractor $F$ and subsequently permuting them using the \textit{feature-space permutation module} $\texttt{permute}$, defined as $f = \texttt{permute}(F(x))$ for each image data $x$, which is similar to the approach proposed in p-\textsc{FeSTA} \cite{9934926} These encrypted features are then transmitted from each client to the server and stored on the server memory, following which all subsequent learning processes are exclusively conducted on the server-side, thus eliminating the need for additional communication and computational overheads on the client-side.

\begin{gather}
\label{eq0}
f = \texttt{permutate}(F(x))
\end{gather}

To ensure encryption of the features, the patch embedder and the position embedding, which are unknown to the server or any outside attacker, are employed in conjunction with the \textit{feature-space permutation module}. As a result, the server is unable to accurately reconstruct the private data from the transmitted feature. Details regarding the formulation and experimental outcomes pertaining to privacy preservation will be discussed in Sections \ref{privacy} and \ref{recons}.


The server employs random masked sampling to conduct SSL, utilizing the encrypted features from all participating clients. This is achieved by modifying the original DINO's local-to-global correspondence learning strategy to small-to-large patch features correspondence, as depicted in Figure~\ref{fig_MS_DINO}. The motivation of the DINO, as illustrated in Figure~\ref{fig_MS_DINO}(A), is to imbue the model with the visual semantics of an image via teacher-student knowledge distillation. Specifically, large crops of \textit{global views} are presented to the teacher, while multiple small crops of \textit{local views} are provided to the student. To adapt this concept, random masked sampling is utilized, replacing the large-sized \textit{global views} by sampling large number of patch features and the smalle-sized of \textit{local views} by sampling smalle number of patch features, as illustrated in Figure~\ref{fig_MS_DINO}(B).
To be specific, for the permuted feature $f$ obtained from an image, the feature $f_{l}$ is created by randomly sampling the majority of permuted patch features, while the feature $f_{s}$ is generated by randomly sampling a relatively smaller number of patches. Similar to DINO, feature $f_{l}$ is fed to the teacher, while both $f_{l}$ and $f_{s}$ are presented to the student. Subsequently, the student is optimized to match the prediction of the momentum teacher by utilizing relatively small information about the image, in accordance with the local-to-global correspondence strategy of DINO.

Assuming that a sampled feature $f_{l}$ is composed of a large number of permuted patch features, and sampled features $f_{s}$ are composed of a small number of permuted patch features, we define the sets of samples containing $M$ differently sampled $f_{l}$ as $V_{l} = {f_{l}^1, ..., f_{l}^M}$, and $N$ differently sampled $f_{s}$ as $V_{s} = {f_{s}^1, ..., f_{s}^N}$. Let the teacher and student models be denoted by $G_{\theta_{t}}$ and $G_{\theta_{s}}$, respectively, and let the cross-entropy loss be denoted by $\mathcal{L}$. The student is trained to mimic the teacher's prediction through the following optimization:
\begin{gather}
\label{eq1}
\min_{G_{\theta_{s}}}{\sum_{f \in V_{l}}\sum_{f_{\prime} \in V_{s}, V_{l}}~\mathcal{L}(G_{\theta_{s}}(f^{\prime}), G_{\theta_{t}}(f))}
\end{gather}
During the learning process, the momentum teacher model is updated with an exponential moving average (EMA) of the student's update, where $\lambda$ follows a cosine scheduling:
\begin{gather}
\label{eq2}
G_{\theta_{t}} = \lambda{G_{\theta_{t}}} + (1 - \lambda)G_{\theta_{s}}
\end{gather}
The algorithm for preparing the encrypted feature set and conducting self-supervised learning with MS-DINO is formally presented in Algorithm~\ref{alg:algo1}. Given the limited data availability for a single client, this pre-trained model may yield improved generalization performance, which is further investigated in Section \ref{ct} and \ref{cxr}.

\begin{algorithm}[!h]
    \SetNoFillComment 
    \caption{Proposed MS-DINO algorithm}
    \label{alg:algo1}
    \DontPrintSemicolon
    \SetKwProg{Fn}{Function}{:}{}
    \SetKwFunction{ServerMain}{ServerMain}
    \SetKwFunction{ClientMain}{ClientMain}
    \SetKwFor{ForP}{for}{do in parallel}{endfor}
        \SetKwFor{For}{for}{}{endfor}
    \SetAlgoNoLine
    \tcc{Run on Client $c$}
    \Fn{\ClientMain}{
        \SetAlgoVlined
        {Client initialize with arbitrary feature embedder $F$}
        
        {
        \For{$\mathbf{data} \ x \in \{1,2,\ldots X\}$}{
        $f = \texttt{permute}(F(x))$ \\
        feature set $\textbf{\textit{f}} =\{f_1, f_2, \ldots f_X\} \leftarrow f$
        }
        \KwRet $\textbf{f}$
        }
    }
 	\BlankLine
    
    \tcc{Run on Main Server}
    \Fn{\ServerMain}{
        \SetAlgoVlined
        Server initializes student $G_{\theta_{s}}$ and teacher $G_{\theta_{t}}$ \\
        \ForP{$\mathbf{clients} \ c \in \{1,2,\ldots C\}$}{
                $ \textbf{\textit{f}}_{c} \leftarrow \ClientMain(c)$ \\
                \texttt{Memory} $= \{\textbf{\textit{f}}_1,\textbf{\textit{f}}_2,\ldots \textbf{\textit{f}}_c\} \leftarrow \textbf{\textit{f}}_c$ \\
                \tcp*[l]{\footnotesize Save all features $\textbf{f}c$ in \texttt{Memory} }
        }
        
        \For{$\mathbf{epoch} \ e \in \{1,2,\ldots E\}$}{
        \For{$\mathbf{features} \ \textbf{f} \in \texttt{\rm \texttt{Memory}}$}{
                \tcp*[l]{\footnotesize Run MS-DINO learning in server}
                $\mathcal{L}_{DINO}={\sum_{f \in V_{l}}\sum_{f_{\prime} \in V_{s}}~\mathcal{L}(G_{\theta_{s}}(f^{\prime}), G_{\theta_{t}}(f))}$
                \BlankLine
                $\theta_{s} \leftarrow \theta_{s} - {\eta \over {N}}{\partial \mathcal{L}_{DINO} \over  {\partial \theta_{s}}}$ \tcp*[l]{\footnotesize update student}
                $\theta_{t} = \lambda{\theta_{t}} + (1 - \lambda)\theta_{s}$ \tcp*[l]{\footnotesize EMA teacher }
                }}
        \ForP{$\mathbf{clients} \ c \in \{1,2,\ldots C\}$}{
        Distribute $G_{\theta}$ to authorized client \\
        \tcp*[l]{\footnotesize Distribute \add{pre-trained} model}
        }
        }
\end{algorithm}

\subsection{Fine-tuning for Tasks of Interest}
The resulting SSL model is accessible to authorized users for specific purposes, as illustrated in Figure~\ref{fig_framework}. For instance, clients can utilize the pre-trained model as a foundation to improve generalization performance when training a model for organ-at-risk (OAR) segmentation in radiotherapy planning, by leveraging its ability to attend to visual semantics within the image.

To provide further clarity, let us consider a client $c$ who employs the self-supervised model backbone, denoted as $G$, and a task-specific layer, such as a decoder, denoted as $H_c$, along with the data and labels for fine-tuning, represented by $x_c$ and $y_c$, respectively. Additionally, let us assume the use of a task-specific loss function denoted as $\mathcal{L}_c$. The optimization problem for fine-tuning can be formulated as follows:
\begin{gather}
\label{eq2}
\min_{G, H}\sum\limits_{i=1}\mathcal{L}_{c}(H_{c}(G(x_{c})), y_{c})
\end{gather}

\begin{figure}[!t]
\centering
\includegraphics[width=0.5\textwidth]{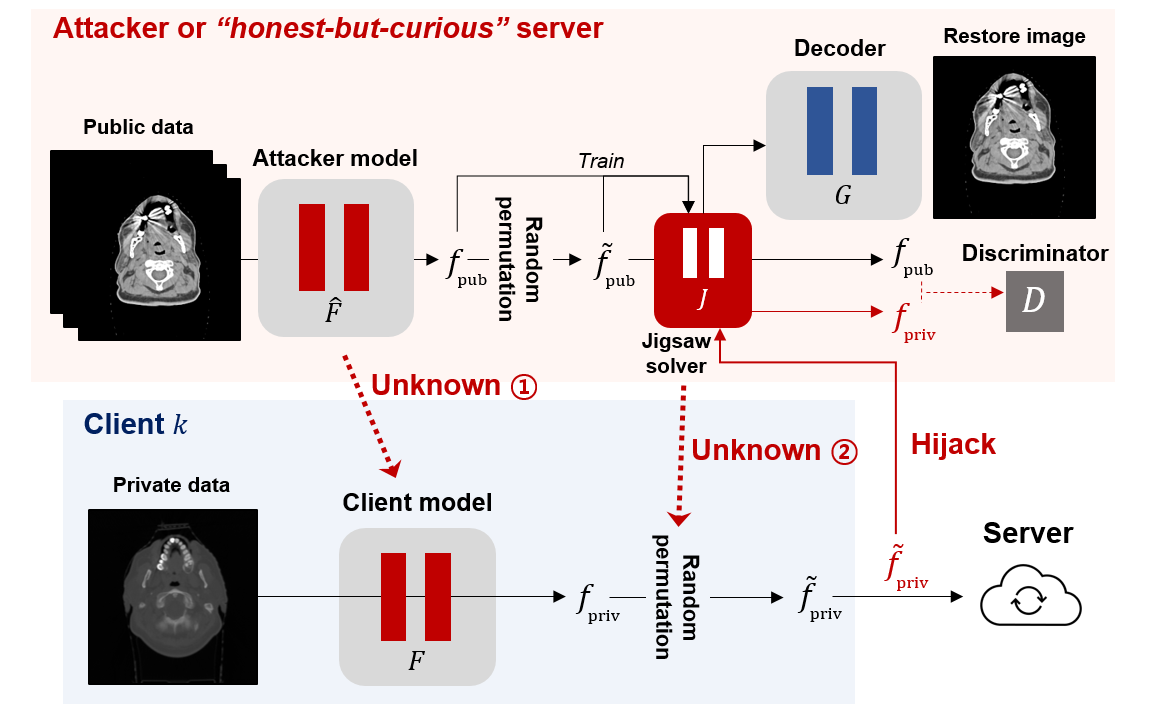}
\caption{Potential attack scenario in the presence of the \textit{feature-space permutation module} involves the attacker having to optimize both the feature-space jigsaw solver and attacker encoder model jointly, which makes it challenging to achieve successful reconstruction of private data.
} 
\label{fig_privacy}
\end{figure}

\subsection{Protecting Privacy with \textit{feature-space permutation module}}
\label{privacy}

The random patch permutation in the feature space has been previously examined as a privacy-preserving technique \cite{9934926}. Since client-side embedded features are kept on the server-side, there is a potential risk of privacy infringement if a "honest-but-curious" server or a malevolent attacker intercepts the features during transmission and endeavors to reconstruct the original data from the features.

Suppose an attacker intercepts the encrypted features with permutation and has adequate access to public data in the same domain. In that case, they would need to solve two problems simultaneously: (1) training an attacker-side feature extractor to embed the image into the feature space identically to the unpermutated version of the encrypted features and (2) training the jigsaw solver to unshuffle the encrypted features into their original order in the feature space of the intercepted features, not in the image space.

To be more precise, let $\hat{F}$ represent the attacker-side model's permutated and original features, embedded as $\Tilde{f}_{pub}$ and $f_{pub}$, respectively. Let $m$ denote the number of attacker-side images, such as publicly available computed tomography (CT) scans, and the hijacked encrypted features as $\Tilde{f}_{priv}$ embedded by an arbitrary client-side feature embedder $F$ and the number of those features as $n$.

The attacker-side model $\hat{F}$, discriminator $D$, and decoder $G$ can be trained by simultaneously optimizing the following two learning objectives:
\begin{align}
\label{eq1}
\min_{\hat{F}}\max_{D}\sum\limits_{i = 1}^{m}\sum\limits_{j = 1}^{n}[\log(1-{D(J(\Tilde{f}^{(i)}_{priv}))})+\log{D(J(\Tilde{f}^{(j)}_{pub}))}]
\end{align}
\begin{gather}
\label{eq2}
\min_{G}\sum\limits_{i=1}^{m}L_{decoder}(G(J(\Tilde{f}^{(i)}_{pub})), x_{pub}^{(i)})
\end{gather}
where $L_{decoder}$ denotes reconstruction loss for decoder.
Meanwhile, the second optimization problem for the jigsaw solve $J$ can be formulated as follow:
\begin{gather}
\label{eq3}
\min_{J}\sum\limits_{i=1}^{m}L_{jigsaw}(J(\Tilde{f}^{(i)}_{pub}), f^{(i)}_{pub})
\end{gather}
where $L_{jigsaw}$ denotes similarity loss in the feature space.



It should be noted that simultaneous solution of the first two equations, Eq.~\eqref{eq1} and Eq.~\eqref{eq2}, the exact jigsaw solver should be unraveled, which can be obtained if the Eq.~\eqref{eq3} is successfully solved. However, the jigsaw solver should be trained and utilized in the same feature space as the hijacked encrypted feature, which necessitates knowing the correct solution for the attacker model $\hat{F}$ to embed the same feature space, and this is conversely the target of the optimization problem Eq.~\eqref{eq1}. Combined, optimization of Eq.~\eqref{eq1}, Eq.~\eqref{eq2}, and Eq.~\eqref{eq3} requires to already have each other's solutions, indicating that the problems are underdetermined and challenging to solve in practical scenarios.

Experimental findings that support this assertion are presented in Section~\ref{recons}.

\section{Results}
\label{result}
We utilized the proposed method in CT imaging and assessed its efficacy on two downstream tasks, namely OAR segmentation and intracranial hemorrhage (ICH) detection.




\subsection{Details of dataset}
To simulate self-supervised distributed learning, we utilized The Cancer Imaging Archive (TCIA) Head-Neck-PET-CT dataset \cite{vallieres2017radiomics}, which consists of CT data obtained from four distinct institutions. We created a collaborative setting where each of the four institutions, namely Centre hospitalier de l’Université de Montréal [CHUM], Centre hospitalier universitaire de Sherbrooke [CHUS], Hôpital Maisonneuve-Rosemont [HMR] de Montréal, and Hôpital général juif [HGJ] de Montréal, participated as an independent client. Table~\ref{tab:data_foundation} provides an overview of these institutions.




\begin{table}[t!]
  \centering
  \caption{Data for MS-DINO self-supervised learning}
          \resizebox{0.5\textwidth}{!}{
    \begin{tabular}{ccccc}
    \toprule
          & \textbf{Client \#1} & \textbf{Client \#2 } & \textbf{Client \#3} & \textbf{Client \#4} \\
    \midrule
    Source & CHUM  & CHUS  & HMR   & HGJ \\
    No. of patient & 21    & 30    & 30    & 30 \\
    No. of slice & 2,873  & 5,388  & 2,906  & 2,674 \\
    \bottomrule
    \end{tabular}%
        }
  \label{tab:data_foundation}%
\end{table}%


In practical implementation, a user, typically from a single institution, usually acquires self-supervised weights and fine-tunes them for a specific task. However, the limited availability of data and labels is a common challenge in such settings. To address this issue, we conducted experiments under two different scenarios: data-abundant (Full) and data-insufficient (Limited). Additionally, we evaluated the generalizability of the proposed SSL approach in two different downstream tasks, namely OAR segmentation and ICH detection.




For the downstream OAR segmentation task, we used the Medical Image Computing and Computer Assisted Intervention (MICCAI) 2015 head and neck challenge dataset \cite{MICCAI} to fine-tune the self-supervised model. The dataset consisted of 2,911 CT slices with labels from 38 patients, where all patients' data were utilized for the data-abundant setting, and 8 patients' data with 608 CT slices were used for the data-insufficient setting.

For the downstream ICH detection task, we used the Computed Tomography Images for Intracranial Hemorrhage Detection and Segmentation Dataset \cite{hssayeni2020computed}, which includes 2,500 brain window CT images from 82 patients. After randomly splitting the test set patients, we used 67 images from 2,343 patients for the data-abundant setting and 15 images from 904 images for the data-insufficient setting.

\begin{table}[t!]
  \centering
  \caption{Data for fine-tuning and evaluate on downstream tasks}
      \resizebox{0.48\textwidth}{!}{
    \begin{tabular}{ccccc}
    \toprule
    \multirow{2}[4]{*}{\textbf{Task}} &       & \multicolumn{2}{c}{\textbf{Amount of data}} & \multirow{2}[4]{*}{\textbf{Test}} \\
\cmidrule{3-4}          &       & \textbf{Full} & \textbf{Limited} &  \\
    \midrule
    \multirow{2}[1]{*}{OAR segmentation} & No. of patient & 38  & 8   & 7 \\
          & No. of slice & 2,911  & 608   & 521 \\
    \multirow{2}[1]{*}{ICH detection} & No. of patient & 67  & 25   & 15 \\
          & No. of slice & 2,343  & 904   & 476 \\
    \bottomrule
    \end{tabular}%
    }
  \label{tab:finetune}%
\end{table}%


For the evaluation of OAR segmentation performance, CT and region-of-interest (ROI) data, which were meticulously collected and delineated by board-certified radiation oncologists from an external institution (Gangnam Severance Hospital), were utilized. Data of 44 head and neck cancer patients who received radiation therapy between 2007 and 2021 were collected, and seven patient data containing all ROIs of head and neck OARs were used for the evaluation.
To evaluate the ICH detection performance, a randomly split subset consisting of 476 images from 15 patients was obtained from the ICH detection dataset.


\subsection{Implementation Details}
The CT images were preprocessed by cropping the central area of size $224 \times 224$ from the original size of $512 \times 512$. 
As the data came from different sources, the pixel spacing was adjusted to match between the datasets. 
We employed the patch embedder of the DINO model, which was pre-trained on ImageNet, as the arbitrary feature embedder. The embedded features were then subjected to permutation using the \textit{feature-space permutation module} proposed in the previous work \cite{9934926}, resulting in encrypted features for the remaining learning process. For the Transformer component, we utilized the transformer of ViT-S with 6 heads, 12 layers, and a patch size of 8. This component was initialized with the self-supervised DINO weights from ImageNet.
As in \cite{caron2021emerging}, we used teacher and student models of identical size. To obtain the global view, we sampled a large number of patch features, leading to a sampling ratio ranging from 0.9 to 1.0. For multiple local views, we sampled a small number of patch features, with a sampling ratio between 0.3 and 0.5. We compared our approach to the original DINO implementation and adopted the same configuration with crop sizes of 0.4-1.0 and 0.05-0.4 for global and local views, respectively. Given the relatively smaller size of the dataset used in the experiment and the reduced complexity of medical images, we opted to decrease the dimension of the DINO head output from 65,536 to 8,192.
For SSL using the proposed MS-DINO method, we employed the Adam W optimizer with a batch size of 8 and a cosine decay scheduler with a maximum learning rate of 0.0001 on the server-side device. The model was trained for 50 epochs. In the case of SSL with the DINO method through FL, we used the same optimizer, scheduler, and learning rate with a batch size of 4 per client. The model was trained for 50 federated rounds on the client-side devices, to match the total number of updates in MS-DINO learning on the server-side. For FL, the server aggregated, averaged, and distributed both the student and teacher model parameters every round.


For the downstream task of OAR segmentation, we employed the ViT backbone and UperNet \cite{xiao2018unified} decoder as the encoder and decoder parts of the segmentation model, following the implementation described in a previous work \cite{he2022masked}. As a baseline for comparison, we also used two recent CNN-based models, R50-UperNet \cite{xiao2018unified} and STDC \cite{fan2021rethinking}. The decoder was designed to perform multi-class segmentation of various ROIs, including the brainstem, optic chiasm, mandible, optic nerves, parotid glands, and submandibular glands. We used the combined Focal loss and Dice loss, following prior work \cite{zhang2021rethinking}, to optimize the model. For the segmentation model, we employed the SGD optimizer with a learning rate of 0.01 and a batch size of 10.
For the downstream task of classification, we added a simple linear layer as the classification head. Similar to segmentation, two recent CNN-based models, ResNext \cite{xie2017aggregated} and ConvNext \cite{liu2022convnet}, were also implemented for comparison.
As we formulated the ICH detection task as binary classification, we optimized the model using the BCE loss. We used the Adam W optimizer with a learning rate of 0.0001 and a batch size of 16 for fine-tuning the model for the classification task.

All experiments were conducted using Python 3.9 and PyTorch 1.10 on NVIDIA RTX 3090. We utilized the FLOWER framework \cite{beutel2020flower} for simulating distributed learning.

\subsection{Details of evaluation}

In order to assess the segmentation performance, the Dice similarity coefficient (DSC) has been utilized to measure the degree of overlap between the predicted segmentation mask and the corresponding ground truth label. To evaluate the detection performance, the area under the receiver operating characteristics curve (AUC) has been computed, and the sensitivity, specificity and accuracy were also calculated.

To statistically compare the results, a non-parametric bootstrap random sampling method was employed, where samples of the same size as the evaluation set were randomly drawn with replacement 1000 times. The confidence intervals were calculated based on the relative frequency distribution of the estimates of these samples, with the interval between the $100 \times (\alpha/2)$ and $100 \times (1 - \alpha/2)$ percentiles used to determine the range.

\subsection{Simulation for Feature Inversion Attack}
We evaluated whether the permutation model is effective to prevent privacy attacks from feature hijacking, supposing the optimal configuration for the malicious attacker. 
In accordance with prior work on encryption using random patch permutation \cite{9934926}, it was assumed that the attacker has gained access to all encrypted features transmitted from all clients and has the exact knowledge of the permutation ratio and patch size, architecture of the unknown patch embedder, and dimensionality of position embedding. To achieve this, the same architecture as the original patch embedder was utilized as the attacker-side feature embedder, and the discriminator and decoder were employed with the three-layered discriminator and the four-layered generator from DCGAN \cite{radford2015unsupervised}, respectively. To solve the random permutation in the feature space, the transformer with 12 encoder layers and 12 attention heads was employed as a jigsaw solver. The discriminator was optimized with a modified version of the GAN loss as formulated in Eqs.~\eqref{eq1} and \eqref{eq2}. The learning objective for the decoder consisted of the combined $L1$ and $L2$ losses, while the objective for the jigsaw solver was $L1$ loss.
Furthermore, it was assumed that the attacker has access to a substantial amount of data in the same domain, which is publicly available. To train the attacker-side networks, we utilized 6,189 CT slices from the TCIA HNSCC-3DCT-RT data \cite{clark2013cancer}.
The model underwent training for a total of 5 epochs, using a batch size of 1 and a learning rate of 0.001.

\begin{figure}[!t]
\centering
\includegraphics[width=0.5\textwidth]{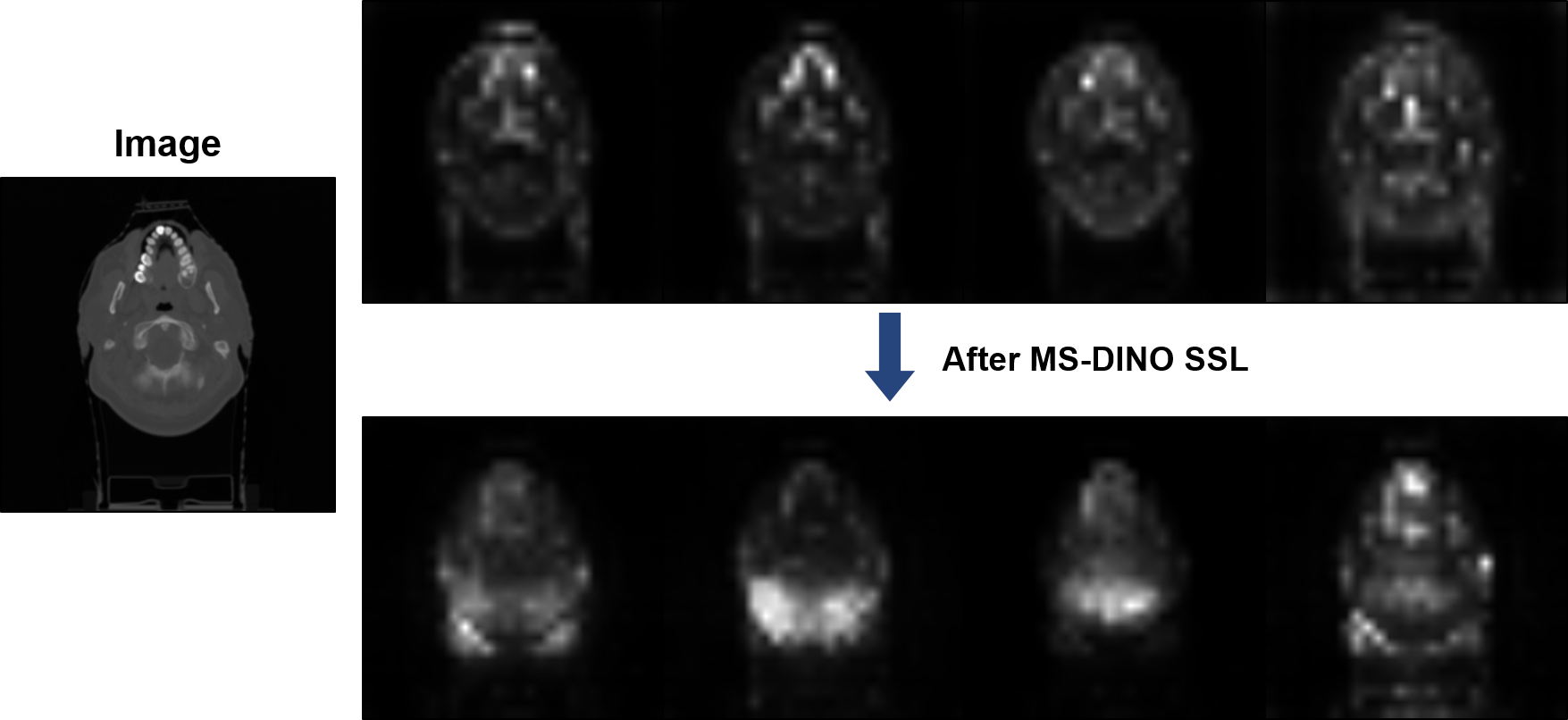}
\caption{The multi-head attention visualization is presented. The obtained attentions after MS-DINO learning were compared with those of the DINO pre-trained ImageNet weights. The attention heads were observed to have become refined and to concentrate on various semantic components. SSL, self-supervised learning.} 
\label{fig_attention}
\end{figure}

\begin{table*}[t!]
  \centering
  \caption{Segmentation results after fine-tuning in data-limited setting}
    \resizebox{1\textwidth}{!}
    {
    \begin{tabular}{ccccccccccc}
    \toprule
    \textbf{Method} & \textbf{Overall } & \textbf{Brainstem} & \textbf{Chiasm} & \textbf{Mandible} & \textbf{Optic n Lt.} & \textbf{Optic n Rt.} & \textbf{Parotid Lt. } & \textbf{Parotid Rt.} & \textbf{SMG Lt.} & \textbf{SMG Rt.} \\
    \midrule
    \textit{\textbf{SOTA-CNN}} &       &       &       &       &       &       &       &       &       &  \\
    R50-UperNet\cite{xiao2018unified} & 0.443 & 0.695 & 0.103 & 0.637 & 0.165 & 0.168 & 0.608 & 0.595 & 0.484 & 0.531 \\
          & (0.405-0.473) & (0.638-0.738) & (0.026-0.198) & (0.600-0.671) & (0.080-0.249) & (0.063-0.286) & (0.553-0.667) & (0.532-0.658) & (0.357-0.594) & (0.484-0.573) \\
    STDC\cite{fan2021rethinking}  & 0.145 & 0.402 & 0.00     & 0.456 & 0.00     & 0.00     & 0.226 & 0.192 & 0.008 & 0.021 \\
          & (0.125-0.165) & (0.363-0.440) & (0.000-0.000) & (0.430-0.482) & (0.000-0.000) & (0.000-0.000) & (0.131-0.329) & (0.103-0.281) & (0.003-0.015) & (0.004-0.039) \\
    \textit{\textbf{ViT-UperNet}} &       &       &       &       &       &       &       &       &       &  \\
    No SSL & 0.427 & 0.621 & 0.082 & 0.609 & 0.191 & 0.053 & 0.618 & 0.600   & 0.508 & 0.565 \\
          & (0.399-0.450) & (0.568-0.666) & (0.007-0.167) & (0.579-0.645) & (0.063-0.321) & (0.012-0.106) & (0.553-0.683) & (0.513-0.678) & (0.468-0.551) & (0.509-0.622) \\
    DC-SSL & \textbf{0.554} & \textbf{0.728} & 0.249 & \textbf{0.739} & 0.296 & \textbf{0.344} & 0.710  & 0.706 & \textbf{0.612} & 0.598 \\
          & \textbf{(0.516-0.579)} & \textbf{(0.650-0.787)} & (0.141-0.345) & \textbf{(0.719-0.759)} & (0.160-0.393) & \textbf{(0.192-0.465)} & (0.664-0.755) & (0.629-0.777) & \textbf{(0.562-0.679)} & (0.495-0.683) \\
    FL-SSL & 0.518 & 0.706 & 0.178 & 0.664 & \textbf{0.327} & 0.303 & 0.660  & 0.680  & 0.547 & 0.599 \\
          & (0.474-0.547) & (0.637-0.757) & (0.058-0.304) & (0.640-0.691) & \textbf{(0.210-0.404)} & (0.175-0.410) & (0.594-0.714) & (0.617-0.726) & (0.488-0.616) & (0.522-0.670) \\
    MS-DINO-SSL & 0.550  & 0.713 & \textbf{0.275} & 0.733 & 0.326 & 0.299 & \textbf{0.723} & \textbf{0.713} & 0.552 & \textbf{0.614} \\
    (proposed) & (0.520-0.575) & (0.622-0.779) & \textbf{(0.162-0.380)} & (0.721-0.745) & (0.207-0.406) & (0.202-0.398) & \textbf{(0.686-0.756)} & \textbf{(0.647-0.765)} & (0.486-0.627) & \textbf{(0.549-0.683)} \\
    \bottomrule
                \multicolumn{10}{l}{\footnotesize  SSL, self-supervised learning; DC, data-centralized; FL, federated learning.}
    \end{tabular}%
    
    }
  \label{tab:CT_limited}%
\end{table*}%

\begin{table*}[t!]
  \centering
  \caption{Segmentation results after fine-tuning in data-abundant setting}
    \resizebox{1\textwidth}{!}
    {
    \begin{tabular}{ccccccccccc}
    \toprule
    \textbf{Method} & \textbf{Overall } & \textbf{Brainstem} & \textbf{Chiasm} & \textbf{Mandible} & \textbf{Optic n Lt.} & \textbf{Optic n Rt.} & \textbf{Parotid Lt. } & \textbf{Parotid Rt.} & \textbf{SMG Lt.} & \textbf{SMG Rt.} \\
    \midrule
    \textit{\textbf{SOTA-CNN}} &       &       &       &       &       &       &       &       &       &  \\
    R50-UperNet\cite{xiao2018unified} & 0.603 & 0.747 & 0.274 & 0.765 & 0.412 & 0.385 & 0.716 & 0.741 & 0.688 & 0.701 \\
          & (0.569-0.628) & (0.691-0.791) & (0.208-0.334) & (0.746-0.783) & (0.362-0.457) & (0.219-0.532) & (0.669-0.765) & (0.688-0.783) & (0.643-0.747) & (0.663-0.733) \\
    STDC\cite{fan2021rethinking}  & 0.474 & 0.652 & 0.158 & 0.596 & 0.242 & 0.267 & 0.682 & 0.625 & 0.529 & 0.514 \\
          & (0.433-0.505) & (0.535-0.744) & (0.062-0.252) & (0.575-0.616) & (0.152-0.311) & (0.147-0.374) & (0.639-0.725) & (0.529-0.704) & (0.451-0.617) & (0.447-0.577) \\
    \textit{\textbf{ViT-UperNet}} &       &       &       &       &       &       &       &       &       &  \\
    No SSL & 0.578 & 0.717 & 0.216 & 0.765 & 0.371 & 0.333 & 0.730  & 0.751 & 0.693 & 0.625 \\
          & (0.541-0.607) & (0.640-0.782) & (0.141-0.294) & (0.748-0.781) & (0.230-0.493) & (0.194-0.464) & (0.690-0.769) & (0.690-0.796) & (0.644-0.748) & (0.564-0.686) \\
    DC-SSL & 0.648 & 0.742 & \textbf{0.319} & 0.806 & 0.455 & 0.460  & \textbf{0.790} & 0.818 & 0.728 & \textbf{0.719} \\
          & (0.617-0.671) & (0.666-0.807) & \textbf{(0.220-0.413)} & (0.798-0.815) & (0.326-0.541) & (0.411-0.520) & \textbf{(0.747-0.830)} & (0.789-0.840) & (0.680-0.779) & \textbf{(0.674-0.760)} \\
    FL-SSL & 0.635 & \textbf{0.768} & 0.280  & 0.806 & 0.456 & 0.467 & 0.770  & 0.800   & 0.700   & 0.667 \\
          & (0.606-0.656) & \textbf{(0.709-0.816)} & (0.205-0.352) & (0.792-0.817) & (0.367-0.520) & (0.400-0.531) & (0.722-0.814) & (0.753-0.831) & (0.651-0.760) & (0.614-0.720) \\
    MS-DINO-SSL & \textbf{0.652} & 0.762 & 0.283 & \textbf{0.825} & \textbf{0.492} & \textbf{0.467} & 0.782 & \textbf{0.821} & \textbf{0.728} & 0.704 \\
    (proposed) & \textbf{(0.632-0.670)} & (0.704-0.811) & (0.205-0.348) & \textbf{(0.813-0.836)} & \textbf{(0.434-0.536)} & \textbf{(0.407-0.530)} & (0.742-0.820) & \textbf{(0.791-0.846)} & \textbf{(0.679-0.786)} & (0.639-0.761) \\
    \bottomrule
                                \multicolumn{10}{l}{\footnotesize  SSL, self-supervised learning; DC, data-centralized; FL, federated learning.}
    \end{tabular}%
    }
  \label{tab:CT_full}%
\end{table*}%

\begin{table}[t!]
  \centering
  \caption{Classification results after fine-tuning in data-limited setting}
      \resizebox{0.48\textwidth}{!}{
    \begin{tabular}{ccccc}
    \toprule
    \textbf{Method} & \textbf{AUC} & \textbf{Sensitivity} & \textbf{Specificity} & \textbf{Accuracy} \\
    \midrule
    \textbf{\textit{SOTA-CNN}} &       &       &       &  \\
    ResNext\cite{xie2017aggregated} & 0.676 & 0.592 & 0.609 & 0.608 \\
          & (0.322-0.819) & (0.000-0.818) & (0.554-0.679) & (0.551-0.674) \\
    ConvNext\cite{liu2022convnet} & 0.444 & 0.604 & 0.306 & 0.322 \\
          & (0.232-0.816) & (0.222-1.000) & (0.120-0.503) & (0.140-0.500) \\
    \textbf{\textit{ViT model}} &       &       &       &  \\
    No SSL & 0.766 & 0.644 & 0.687 & 0.684 \\
          & (0.622-0.885) & (0.278-1.000) & (0.600-0.774) & (0.602-0.757) \\
    DC-SSL & 0.866 & 0.703 & 0.771 & 0.768 \\
          & (0.766-0.945) & (0.357-0.938) & (0.721-0.825) & (0.717-0.825) \\
    FL-SSL & 0.769 & 0.618 & 0.701 & 0.697 \\
          & (0.466-0.878) & (000-0.818) & (0.628-0.779) & (0.625-0.769) \\
    MS-DINO-SSL & \textbf{0.903} & \textbf{0.795} & \textbf{0.862} & \textbf{0.859} \\
    (proposed) & \textbf{(0.792-0.954)} & \textbf{(0.500-1.000)} & \textbf{(0.800-0.920)} & \textbf{(0.801-0.913)} \\
    \bottomrule
                                \multicolumn{5}{l}{\scriptsize  SSL, self-supervised learning; DC, data-centralized; FL, federated learning. 
                                }
                                
    \end{tabular}%
    }
  \label{tab:CXR_limited}%
\end{table}%

\subsection{Attention Changes with MS-DINO learning}
Figure \ref{fig_attention} illustrates changes in the last layer of multi-head attentions within the ViT model. Prior to SSL with MS-DINO, the attentions of the ViT model pretrained on ImageNet were dispersed throughout the image (upper) with minimal differences in attention between individual heads, suggesting that many attention heads were processing the image redunduntly. However, after performing the MS-DINO SSL, different heads began attending to various semantic components (lower), potentially leading to improved performance with diverse self-attention patterns.

\begin{table}[t!]
  \centering
  \caption{Classification results after fine-tuning in data-abundant setting}
      \resizebox{0.48\textwidth}{!}{
    \begin{tabular}{ccccc}
    \toprule
    \textbf{Method} & \textbf{AUC} & \textbf{Sensitivity} & \textbf{Specificity} & \textbf{Accuracy} \\
    \midrule
    \textbf{\textit{SOTA-CNN}} &       &       &       &  \\
    ResNext\cite{xie2017aggregated} & 0.767 & 0.575 & 0.713 & 0.706 \\
          & (0.474-0.881) & (0.000-0.800) & (0.634-0.799) & (0.628-0.787) \\
    ConvNext\cite{liu2022convnet} & 0.705 & 0.656 & 0.621 & 0.622 \\
          & (0.486-0.899) & (0.222-1.000) & (0.541-0.715) & (0.546-0.702) \\
    \textbf{\textit{ViT model}} &       &       &       &  \\
    No SSL & 0.812 & 0.796 & 0.640  & 0.650 \\
          & (0.650-0.905) & (0.500-0.909) & (0.583-0.713) & (0586-0.726) \\
    DC-SSL & \textbf{0.938} & 0.718 & \textbf{0.918} & \textbf{0.907} \\
          & \textbf{(0.887-0.975)} & (0.500-0.875) & \textbf{(0.873-0.960)} & \textbf{(0.860-0.949)} \\
    FL-SSL & 0.869 & 0.760  & 0.819 & 0.816 \\
          & (0.726-0.940) & (0.500-1.000) & (0.731-0.896) & (0.735-0.887) \\
    MS-DINO-SSL & 0.933 & \textbf{0.846} & 0.903 & 0.900 \\
    (proposed) & (0.778-0.984) & \textbf{(0.500-1.000)} & (0.848-0.950) & (0.852-0.944) \\
    \bottomrule
                                \multicolumn{5}{l}{\scriptsize  SSL, self-supervised learning; DC, data-centralized; FL, federated learning. 
                                }
    \end{tabular}%
    }
  \label{tab:CXR_full}%
\end{table}%

\subsection{Performance Comparison for downstream OAR segmentation}
\label{ct}
\begin{figure}[!t]
\centering
\includegraphics[width=0.45\textwidth]{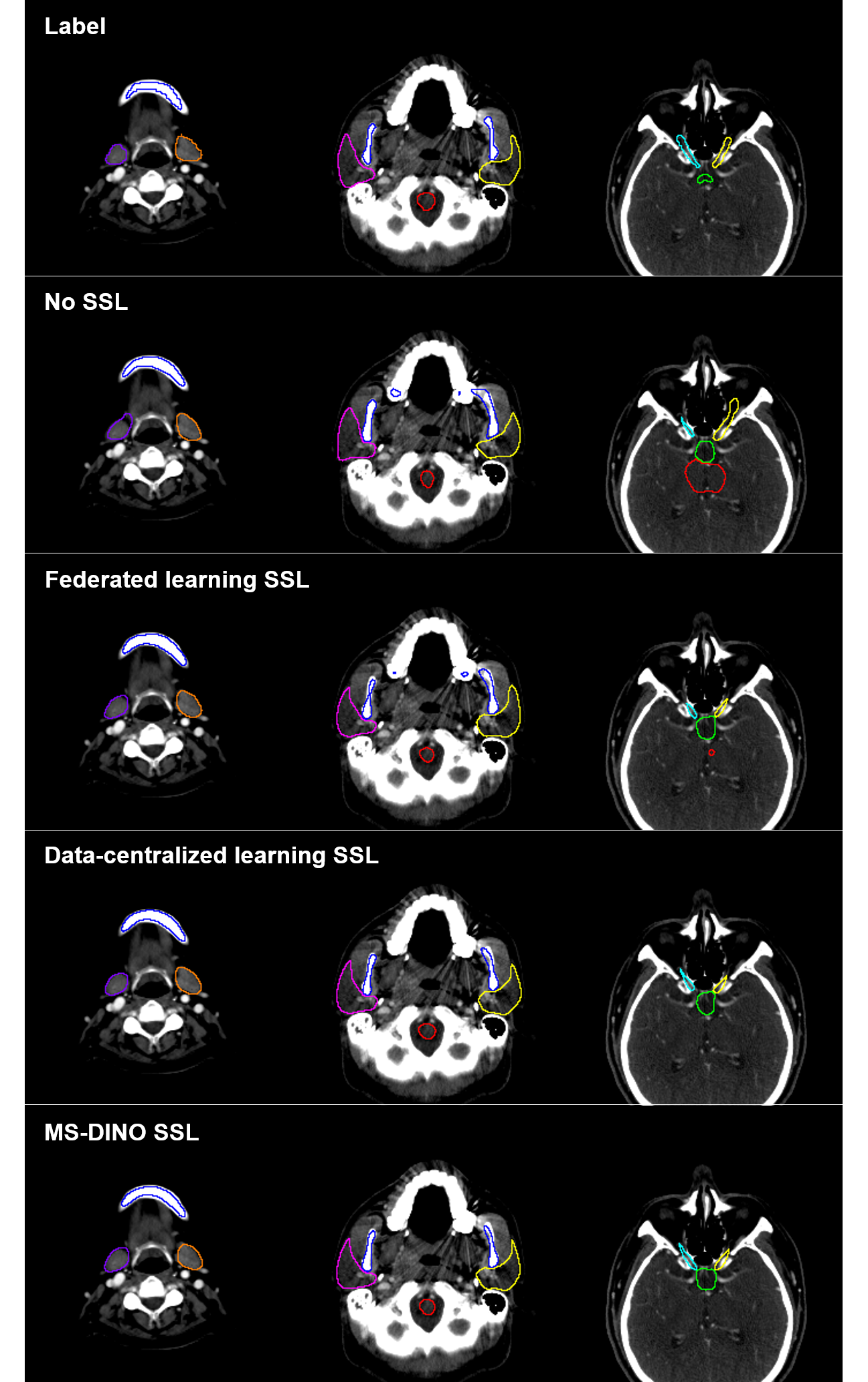}
\caption{Qualitative comparisons of the segmentation results. The model fine-tuned from the pre-trained model using the MS-DINO method displayed comparable results to those obtained with the data-centralized DINO method while outperforming the no self-supervised learning (SSL) baseline. These results were obtained from fine-tuning in a data-abundant setting.} 
\label{fig_segmentation}
\end{figure}

Table~\ref{tab:CT_limited} and \ref{tab:CT_full} present the comparison results for the OAR segmentation task among the methods. The model fine-tuned from the self-supervised weights demonstrated the performance superior to models trained from scratch. Notably, the self-supervised model obtained using the proposed MS-DINO method performed on par with the model obtained from data-centralized SSL, while also outperforming the model obtained using FL.
Figure \ref{fig_segmentation} presents a qualitative comparison among different methods. Specifically, the model fine-tuned from the SSL using the proposed MS-DINO method yielded more accurate predictions for the OAR areas compared to the baseline model that was only fine-tuned. Moreover, the performance of this model was comparable to that of the self-supervised and fine-tuned model trained in a data-centralized manner, and outperformed the those with FL.



\subsection{Performance comparison for downstream ICH detection}
\label{cxr}
Table~\ref{tab:CXR_limited} and \ref{tab:CXR_full} present the comparison results of different methods for the ICH detection task. The results show that models fine-tuned from self-supervised weights generally outperformed the fine-tune-only baselines that were trained from scratch. Furthermore, the self-supervised model obtained with the proposed MS-DINO method demonstrated comparable performance to that of a data-centralized learning, and outperformed that of FL among the self-supervised models. Notably, the proposed method demonstrated a more remarkable benefit in the data-limited setting, highlighting its potential to improve performance with limited data.



\subsection{Communication costs}
When specifying the number of data as $D$, the total number of training rounds as $R$, the round between aggregation and distribution as $r$, the model parameter as $P$, and the size of encrypted feature for each data as $F$, the total communication costs $T$ for SSL with FL and MS-DINO can be expressed as follows:
\begin{gather}
   T_\text{FL} =  {4R \over r} \times P, \\
   T_\text{MS-DINO} = {D \times F + P}
\end{gather}
where the constant 4 is multiplied to account for the both-way parameter transmission of both the teacher and student models, and the both-way transmissions between server and client during FL.
The MS-DINO method proposed in this study does not necessitate continuous communication. Instead, only a single-round communication occurs at the outset, which involves the transmission of encrypted features between the server and clients, and the subsequent model download by the client at the end of the learning process. This approach results in a significant reduction in communication overheads, as illustrated in Figure~\ref{fig_MS_DINO}(A)-(B).


Table~\ref{tab:communication} presents the numerical comparison results. The proposed MS-DINO method requires a communication cost of about only 25\% of that required during the FL. Moreover, this advantage can be further amplified as the number of total learning rounds (epochs) increases.

\subsection{Effects of \textit{feature-space permutation module}}
\label{recons}

Given that the \textit{feature-space permutation module} is one of the key component of the proposed method, we performed the experiments to examine its effects on the privacy protection and the downstream performances.

\subsubsection{Effect on privacy protection}
\begin{figure}[!t]
\centering
\includegraphics[width=0.4\textwidth]{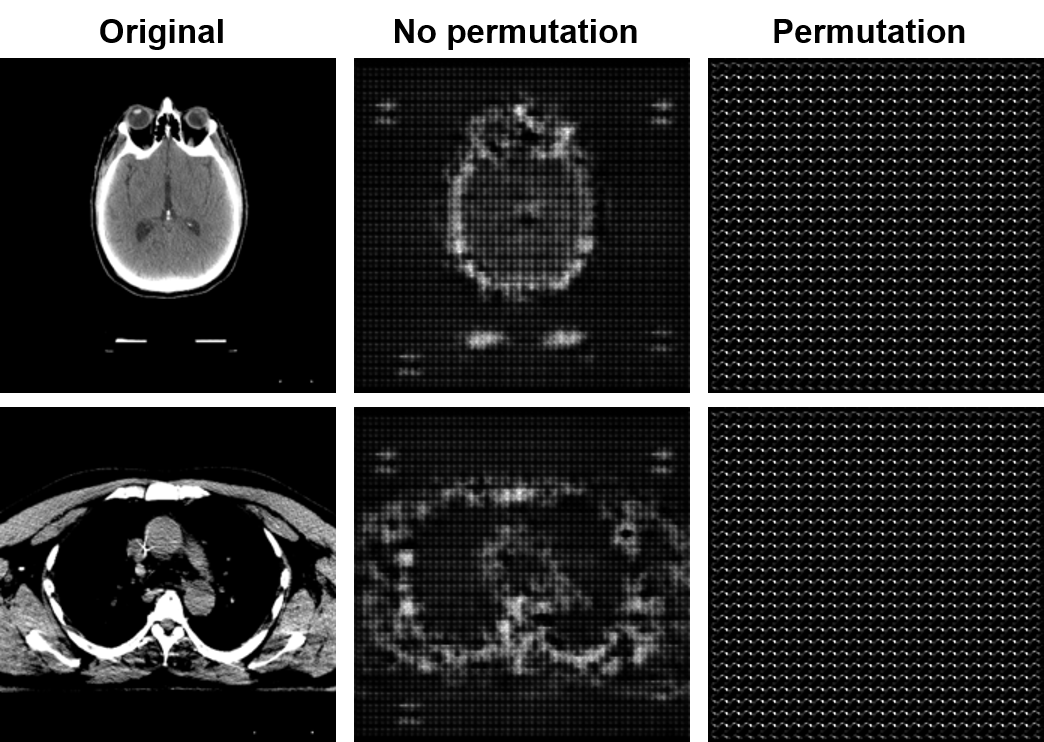}
\caption{Reconstruction results of the privacy attack. Without the \textit{feature-space permutation module}, some degree of the information (level of CT slice, presence of disease, etc.) can be inferred from the reconstructed results, while it was infeasible when the \textit{\textit{feature-space permutation module}} was utilized.} 
\label{fig_recon}
\end{figure}

Table~\ref{tab:recon} and Fig.~\ref{fig_recon} present the qualitative and quantitative analysis results of the reconstruction from the privacy attack to examine the role of the \textit{\textit{feature-space permutation module}} on the privacy protection and performances. The results show that, without the \textit{feature-space permutation module} and with the optimal configuration for the attacker, the private data can be reconstructed to an extent that the privacy attributes such as shape, anatomic location, and disease status of the subject can be inferred. This suggests that the unknown feature embedder can be approximately solved by the attacker. However, with the \textit{\textit{feature-space permutation module}}, it is almost impossible to reconstruct the data to the extent that the privacy can be inferred. These results confirm the claim that solving two optimization problems simultaneously, which need each other's solution, can be considered an underdetermined problem that is practically difficult to solve.

\begin{table}[t!]
  \centering
  \caption{Comparison of communication overheads between methods}
          \resizebox{0.5\textwidth}{!}{
    \begin{tabular}{ccccc}
    \toprule
    \multirow{2}[4]{*}{\textbf{Methods}} & \multicolumn{1}{c}{\multirow{2}[4]{*}{\textbf{No. rounds}}} & \multicolumn{3}{c}{\textbf{Communication overheads}} \\
\cmidrule{3-5}          &       & \textbf{Total} & \textbf{Feature} & \textbf{Weight} \\
    \midrule
    MS-DINO learning & 1     & 1,086.3 M  & 1,043.0 M  & 43.3 M  \\
    Federated learning & 50    & 4,334.1 M  & -     & 2,600.4 M  \\
    \bottomrule
    \end{tabular}%
    }
  \label{tab:communication}%
\end{table}%

\subsubsection{Effect on performances}

Table~\ref{tab:permutation} presents the segmentation performance of the OAR segmentation and ICH detection with and without the \textit{feature-space permutation module}. The results reveal that there is no statistically significant difference in the performance of both tasks according to the use of the \textit{feature-space permutation module}.
These findings provide experimental evidence supporting the patch permutation-invariant property of ViT.

\begin{table}[t!]
  \centering
  \caption{Effect of permutation module on privacy protection}
          \resizebox{0.4\textwidth}{!}{
    \begin{tabular}{ccc}
    \toprule
    \multirow{2}[4]{*}{\textbf{Permutation module}} & \multicolumn{2}{c}{\textbf{Metrics}}\\
\cmidrule{2-3}          & \textbf{MSE} & \textbf{SSIM}\\
    \midrule
    \xmark & 0.237 (0.005) & 0.694 (0.008) \\
    \cmark & \textbf{0.272 (0.009)} & \textbf{0.479 (0.038)} \\
    \bottomrule
                                        \multicolumn{3}{l}{\scriptsize MSE, mean squared error; SSIM, structural similarity index}
    \end{tabular}%
    }
  \label{tab:recon}%
\end{table}%


\begin{table}[t!]
  \centering
  \caption{Effect of permutation module on performances}
        \resizebox{0.42\textwidth}{!}{
    \begin{tabular}{ccc}
    \toprule
    \multirow{2}[2]{*}{} & \textbf{OAR segmentation} & \textbf{ICH detection} \\
        \cmidrule{2-3}   & \textbf{Overall DSC} & \textbf{AUC} \\
    \midrule
    \textbf{\textit{Limited}} &       &  \\
    \multirow{2}[0]{*}{Permutation} & \textbf{0.550}  & \textbf{0.903} \\
          & \textbf{(0.520-0.575)} & \textbf{(0.792-0.954)} \\
    \multirow{2}[0]{*}{No permutation} & 0.539 & 0.889 \\
          & (0.512-0.562) & (0.752-0.952) \\
    \textbf{\textit{Full}} &       &  \\
    \multirow{2}[0]{*}{Permutation} & \textbf{0.652} & 0.933 \\
          & \textbf{(0.632-0.670)} & (0.778-0.984) \\
    \multirow{2}[1]{*}{No permutation} & 0.647 & \textbf{0.949} \\
          & (0.620-0.667) & \textbf{(0.880-0.984)} \\
              \bottomrule
                        \multicolumn{3}{l}{\scriptsize OAR, organ-at-risk; ICH, intracranial hemorrhage.} \\
                        \multicolumn{3}{l}{\scriptsize DSC, dice similairty coefficient; AUC, area under the curve.}
    \end{tabular}%
    }
  \label{tab:permutation}%
\end{table}%

\section{Discussion}
Deep learning-based vision models have exhibited impressive performance; however, their data-driven learning paradigm poses practical challenges for developing AI models for healthcare research, where training data may contain sensitive personal information \cite{perone2019promises}. Additionally, label dependency presents another challenge as labels annotated by medical experts are expensive and difficult to obtain.


To address these challenges, distributed learning methods have been introduced to enable model training without directly sharing private data, and SSL methods have been investigated to alleviate label dependency. However, combining these techniques has led to suboptimal performance compared to data-centralized approaches \cite{makhija2022federated}. Furthermore, FL has inherent properties, namely model aggregation, averaging and distribution by the central server, that multiple rounds communication between server and clients, potentially compromising privacy by enabling attacks by malicious attackers \cite{10025466}.

To address the aforementioned issues, we propose MS-DINO, a novel SSL method that can be used in a distributed manner only with single-round communication while achieving performance comparable to data-centralized SSL. We leverage the permutation-invariant property of the self-attention and the semantic learning via local-to-global correspondence using the teacher-student knowledge distillation, inspired by the intriguing properties of the ViT \cite{naseer2021intriguing}. Specifically, we replace the multi-crop strategy with a random masked sampling strategy, consistent with pioneering SSL approaches that use masked image modeling with ViT in a patch-wise manner \cite{bao2021beit, he2022masked}. Additionally, we incorporate the random \textit{feature-space permutation module} \cite{9934926} to enhance privacy as well as enabling single-round self-supervised distributed learning without compromising the performance. This enables encrypted patch features to be storted in the server-side memory, allowing SSL based on masked random sampling of patch features to be performed on the server-side device while preserving the privacy of participating subjects. 
Previous works \cite{kawamura2020privacy, ivan2019convolutional, sharma2018image} have investigated the use of permutation in a patch or pixel-level for data encryption. However, these works apply permutation to an image itself, which can relatively easily be solved with the model like a jigsaw solver if sufficient data with similar properties are available \cite{noroozi2016unsupervised}. In contrast, our method permutates the patch and position embedded features in the \textit{feature space}, rendering it more challenging to solve. Moreover, little research has examined the use of random patch permutation in conjunction with the permutation-invariant property of the ViT.
Recently, several studies have proposed the methods that combines ViT with the SSL with MAE under FL \cite{wu2022federated, yan2023label}. However, the SSL with MAE requires the direct calculation of error between the predicted and original pixel values, making it infeasible to apply the \textit{feature-space random patch permutation} for encryption. Therefore, these studies simply combine MAE and FL by training the ViT model on the client-side device with MAE SSL and by aggregating the weights with the existing FL method, unlike our approach that leverages the unique properties of ViT.

Our study has several limitations. Firstly, the proposed MS-DINO method has a dependency on the intrinsic permutation-invariant property of the self-attention of Transformer, which limits its use to models that do not have this property, such as the CNN-based architectures. Secondly, while the study has demonstrated that the \textit{feature-space patch permutation} enhances privacy, it is still prone to other malicious attacks in distributed learning, such as model or data poisoning \cite{lyu2020threats, tolpegin2020data, bagdasaryan2020backdoor}, and membership inference attacks \cite{gupta2021membership, shokri2017membership, nasr2019comprehensive, zou2020privacy}, which fall outside the scope of this research. The proposed method can incorporate additional techniques for privacy protection \cite{lyu2020privacy}. Thirdly, the study has not accounted for practical considerations such as the data imbalance or the straggler problem \cite{li2020federated}. Lastly, a direct comparison between the proposed MS-DINO method and the existing ViT-based distributed learning methods, such as \textsc{FeSTA} and \textit{p-\textsc{FeSTA}}, was not possible as they are designed for supervised learning, whereas the proposed method is designed for the SSL for task-agnostic self-supervised model.

\section{Conclusion}
Given the limited data and label availability and the importance of privacy in health research, our method has great promise to enable SSL with only single-round distributed learning in privacy protecting way. Since the self-supervised model obtained with the proposed MS-DINO has a general understanding of visual semantics, it can be used as the task-agnostic self-supervised model to enhance the performances of any downstream task, suggesting its widespread applicability in medical imaging.

\section*{Acknowledgement}
This research was supported in part by a grant of the MD-PhD/Medical Scientist Training Program through the Korea Health Industry Development Institute (KHIDI), funded by the Ministry of Health \& Welfare, Republic of Korea, by the National Research Foundation of Korea (NRF) grant funded by the Korean Government Ministry of Science and ICT (NRF-2020R1A2C1102559), by a faculty research grant of Yonsei University College of Medicine (6-2019-0071), by the National Research Foundation of Korea under Grant NRF-2020R1A2B5B03001980, and by the KAIST Key Research Institute (Interdisciplinary Research Group) Project.

\bibliographystyle{IEEEtran}
\bibliography{fed}

\end{document}